\documentclass[10pt,twocolumn,letterpaper]{article}
\usepackage{iccv}
\usepackage{times}
\usepackage{epsfig}
\usepackage{graphicx}
\usepackage{multicol}
\usepackage{tabularx, booktabs}
\usepackage{amsmath}
\usepackage{amssymb}
\usepackage{pifont}
\usepackage{color}
\usepackage{threeparttable}
\newcommand{\red}[1]{\textcolor{red}{#1}}

\newcommand{\ignore}[1]{}
\usepackage[font=small]{caption}  
\usepackage{microtype}
\newcolumntype{x}[1]{>{\centering\let\newline\\\arraybackslash\hspace{0pt}}p{#1}}
\usepackage[breaklinks=true,bookmarks=false]{hyperref}

\iccvfinalcopy 

\def\iccvPaperID{3151} 

\title{Point-Based Multi-View Stereo Network}

\author{Rui Chen$^{1,3}$\textsuperscript{*} \qquad Songfang Han$^{2,3}$\textsuperscript{*} \qquad Jing Xu$^{1}$ \qquad Hao Su$^{3}$ \vspace{1pt}\\
$^{1}$Tsinghua University  \quad $^{2}$The Hong Kong University of Science and Technology \\ $^{3}$University of California, San Diego \\
{\tt\small chenr17@mails.tsinghua.edu.cn} \quad {\tt\small shanaf@connect.ust.hk} \\ 
{\tt\small jingxu@tsinghua.edu.cn} \quad \quad \quad  {\tt\small \quad haosu@eng.ucsd.edu}\\
}

\makeatletter
\def\blfootnote{\xdef\@thefnmark{}\@footnotetext}
\makeatother

\begin{document}

\twocolumn[{%
\renewcommand\twocolumn[1][]{#1}%
\maketitle
\vspace{-0.4cm}
\begin{center}
    \setlength{\abovecaptionskip}{0cm}
    \centering
    \vspace{-0.4cm}
    \includegraphics[width=0.8\textwidth]{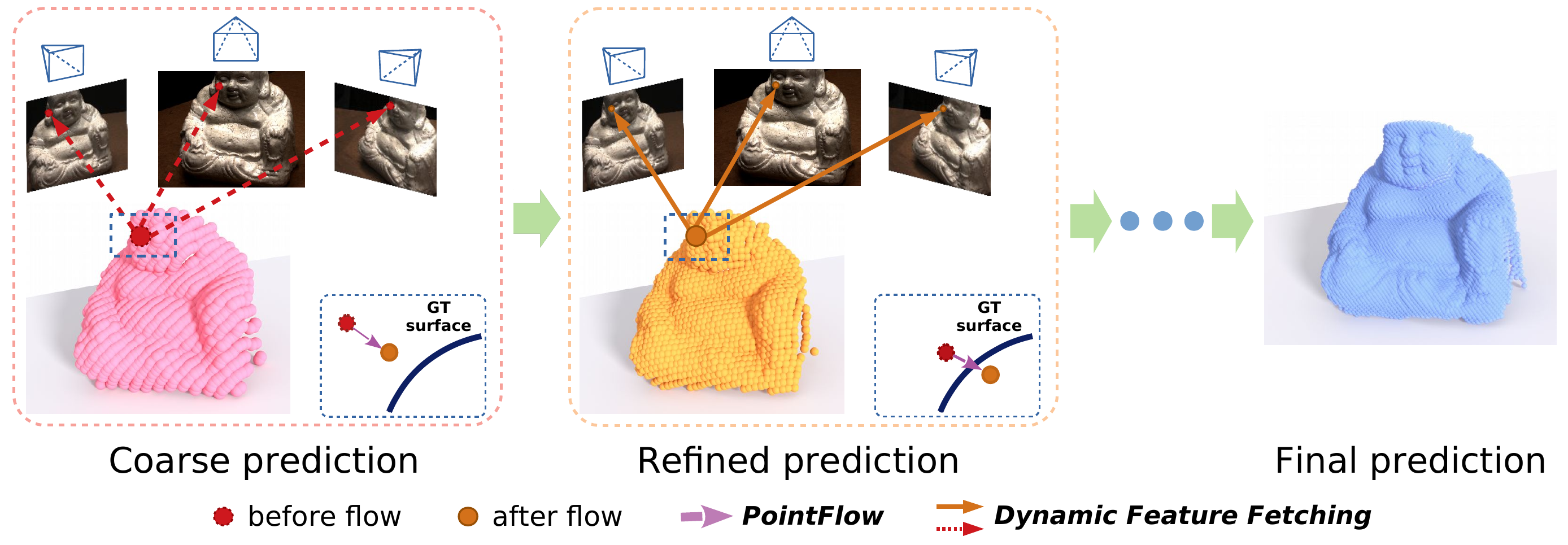}
    \captionof{figure}{Point-MVSNet performs multi-view stereo reconstruction in a coarse-to-fine fashion, learning to predict the 3D flow of each point to the groundtruth surface based on geometry priors and 2D image appearance cues dynamically fetched from multi-view images and regress accurate and dense point clouds iteratively.}
    \label{fig:teaser}
\end{center}%
}]

\blfootnote{\textsuperscript{*} Equal contribution.}

\begin{abstract}
We introduce Point-MVSNet, a novel point-based deep framework for multi-view stereo (MVS). Distinct from existing cost volume approaches, our method directly processes the target scene as point clouds. More specifically, our method predicts the depth in a coarse-to-fine manner. We first generate a coarse depth map, convert it into a point cloud and refine the point cloud iteratively by estimating the residual between the depth of the current iteration and that of the ground truth. Our network leverages 3D geometry priors and 2D texture information jointly and effectively by fusing them into a feature-augmented point cloud, and processes the point cloud to estimate the 3D flow for each point. This point-based architecture allows higher accuracy, more computational efficiency and more flexibility than cost-volume-based counterparts.  Experimental results show that our approach achieves a significant improvement in reconstruction quality compared with state-of-the-art methods on the DTU and the Tanks and Temples dataset. Our source code and trained models are available at \hyperlink{https://github.com/callmeray/PointMVSNet}{https://github.com/callmeray/PointMVSNet}.

\end{abstract}


\section{Introduction}

Recent learning-based multi-view stereo (MVS) methods~\cite{ji2017surfacenet_an-1, yao2018mvsnet_depth-1, huang2018deepmvs_learning} have shown great success compared with their traditional counterparts as learning-based approaches are able to learn to take advantage of scene global semantic information, including object materials, specularity, and environmental illumination, to get more robust matching and more complete reconstruction. All these approaches apply dense multi-scale 3D CNNs to predict the depth map or voxel occupancy. However, 3D CNNs require memory cubic to the model resolution, which can be potentially prohibitive to achieving optimal performance. While Maxim et al.~\cite{tatarchenko2017octreegen} addressed this problem by progressively generating an Octree structure, the quantization artifacts brought by grid partitioning still remain, and errors may accumulate since the tree is generated layer by layer. 

In this work, we propose a novel point cloud multi-view stereo network, where the target scene is directly processed as a point cloud, a more efficient representation, particularly when the 3D resolution is high. Our framework is composed of two steps: first, in order to carve out the approximate object surface from the whole scene, an initial coarse depth map is generated by a relatively small 3D cost volume and then converted to a point cloud. Subsequently, our novel \textit{PointFlow} module is applied to iteratively regress accurate and dense point clouds from the initial point cloud. Similar to ResNet~\cite{he2016resnet}, we explicitly formulate the \textit{PointFlow} to predict the residual between the depth of the current iteration and that of the ground truth. The 3D flow is estimated based on geometry priors inferred from the predicted point cloud and the 2D image appearance cues dynamically fetched from multi-view input images (\autoref{fig:teaser}). 

We find that our Point-based Multi-view Stereo Network (Point-MVSNet) framework enjoys advantages in accuracy, efficiency, and flexibility when it is compared with previous MVS methods that are built upon a predefined 3D volume with the fixed resolution to aggregate information from views. Our method adaptively samples potential surface points in the 3D space. It keeps the continuity of the surface structure naturally, which is necessary for high precision reconstruction. 
Furthermore, because our network only processes valid information near the object surface instead of the whole 3D space as is the case in 3D CNNs, the computation is much more efficient. Lastly, the adaptive refinement scheme allows us to first peek at the scene at coarse resolution and then densify the reconstructed point cloud only in the region of interest. For scenarios such as interaction-oriented robot vision, this flexibility would result in saving of computational power.

\begin{figure*}[t!]
\setlength{\abovecaptionskip}{0cm}
\setlength{\belowcaptionskip}{-0.3cm}
    \centering
    \includegraphics[width=0.9\textwidth]{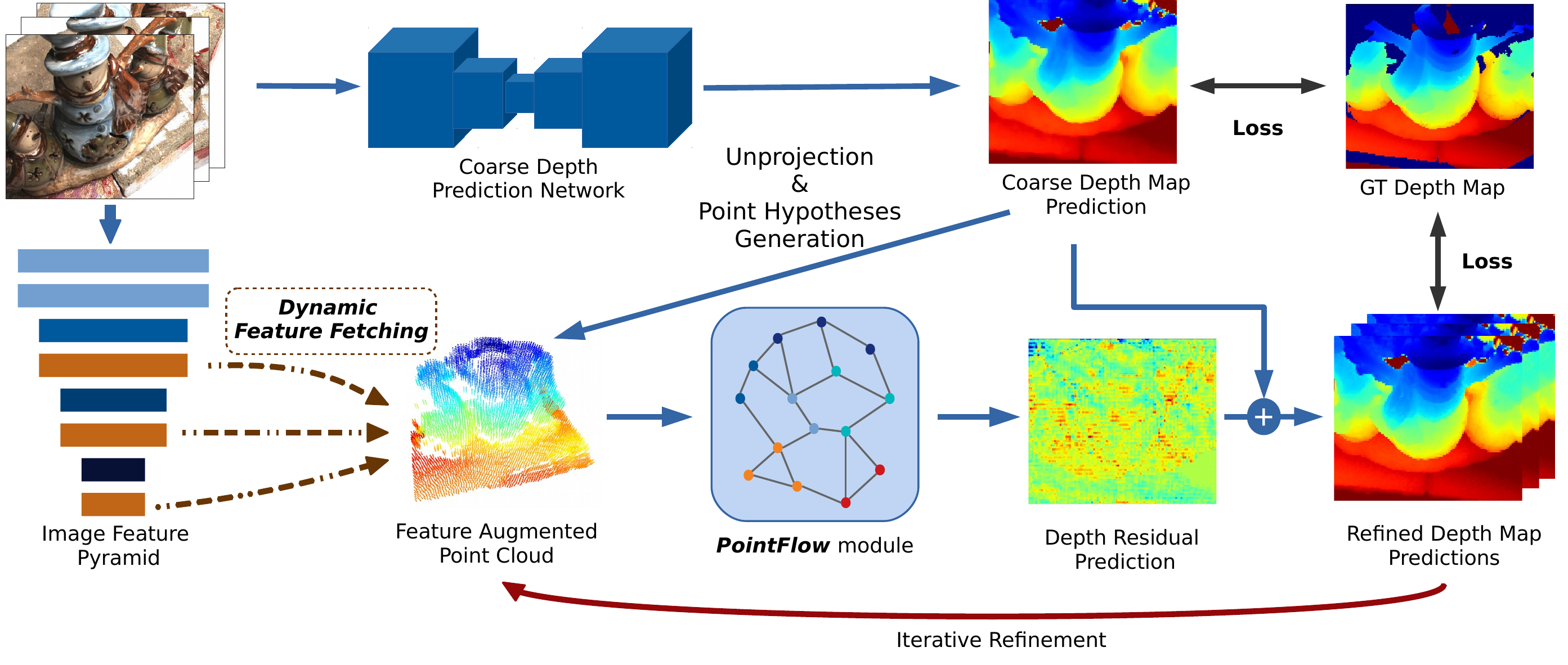}
    \caption{Overview of Point-MVSNet architecture.  A coarse depth map is first predicted with low GPU memory and computation cost and then unprojected to a point cloud along with hypothesized points. For each point, the feature is fetched from the multi-view image feature pyramid dynamically. The \textit{PointFlow} module uses the feature augmented point cloud for depth residual prediction, and the depth map is refined iteratively. }
    \label{fig:whole_architecture}
\end{figure*}

Our method achieves state-of-the-art performance on standard multi-view stereo benchmarks among learning-based methods, including DTU~\cite{aanaes2016LargeScaleDataMultipleView} and Tanks and Temples~\cite{knapitsch2017tanks_and}. Compared with previous state-of-the-art, our method produces better results in terms of both completeness and overall quality. Besides, we show potential applications of our proposed method, such as foveated depth inference.

\section{Related work}
\paragraph{Multi-view Stereo Reconstruction}
MVS is a classical problem that had been extensively studied before the rise of deep learning. A number of 3D representations are adopted, including volumes~\cite{vogiatzis2007multiview, hornung2006hierarchical}, deformation models~\cite{esteban2004silhouette, zaharescu2007transformesh}, and patches~\cite{furukawa2010accurate_dense}, which are iteratively updated through multi-view photometric consistency and regularization optimization. Our iterative refinement procedure shares a similar idea with these classical solutions by updating the depth map iteratively. However, our learning-based algorithm achieves improved robustness to input image corruption and avoids the tedious manual hyper-parameters tuning.

\paragraph{Learning-based MVS}
Inspired by the recent success of deep learning in image recognition tasks, researchers began to apply learning techniques to stereo reconstruction tasks for better patch representation and matching~\cite{han2015matchnet, seki2017sgm, knobelreiter2017cnncrf}. Although these methods in which only 2D networks are used have made a great improvement on stereo tasks, it is difficult to extend them to multi-view stereo tasks, and their performance is limited in challenging scenes due to the lack of contextual geometry knowledge. Concurrently, 3D cost volume regularization approaches have been proposed~\cite{kendall2017gcnet, ji2017surfacenet_an-1, kar2017lsm},  where a 3D cost volume is built either in the camera frustum or the scene. Next, the 2D image features of multi-views are warped in the cost volume, so that 3D CNNs can be applied to it. The key advantage of 3D cost volume is that the 3D geometry of the scene can be captured by the network explicitly, and the photometric matching can be performed in 3D space, alleviating the influence of image distortion caused by perspective transformation and potential occlusions, which makes these methods achieve better results than 2D learning based methods. Instead of using voxel grids, in this paper we propose to use a point-based network for MVS tasks to take advantage of 3D geometry learning without being buredened by the inefficiency found in 3D CNN computation.

\paragraph{High-Resolution MVS}
High-resolution MVS is critical to real applications such as robot manipulation and augmented reality. Traditional methods ~\cite{lhuillier2005quasi, furukawa2010accurate_dense, owens2013shape_anchors} generate dense 3D patches by expanding from confident matching key-points repeatedly, which is potentially time-consuming. These methods are also sensitive to noise and change of viewpoint owing to the usage of hand-crafted features. Recent learning methods try to ease memory consumption by advanced space partitioning~\cite{riegler2017octnet, Wang-2017-OCNN, tatarchenko2017octreegen}. However, most of these methods construct a fixed cost volume representation for the whole scene, lacking flexibility. In our work, we use point clouds as the representation of the scene, which is more flexible and enables us to approach the accurate position progressively.

\paragraph{Point-based 3D Learning}
Recently, a new type of deep network architecture has been proposed in ~\cite{qi2016PointNetDeepLearning, qi2017PointNetDeepHierarchical}, which is able to process point clouds directly without converting them to volumetric grids. Compared with voxel-based methods, this kind of architecture concentrates on the point cloud data and saves unnecessary computation. Also, the continuity of space is preserved during the process. While PointNets have shown significant performance and efficiency improvement in various 3D understanding tasks, such as object classification and detection~\cite{qi2017PointNetDeepHierarchical}, it is under exploration how this architecture can be used for MVS task, where the 3D scene is unknown to the network. In this paper, we propose \textit{PointFlow} module, which estimates the 3D flow based on joint 2D-3D features of \textit{point hypotheses}.

\section{Method}
This section describes the detailed network architecture of Point-MVSNet (\autoref{fig:whole_architecture}). Our method can be divided into two steps, coarse depth prediction, and iterative depth refinement. Let ${\mathbf{I}_0}$ denote the reference image and ${\left\{\mathbf{I}_i\right\}}_{i=1}^N$ denote a set of its neighbouring source images. We first generate a coarse depth map for ${\mathbf{I}_0}$ (\autoref{sec:coarse_depth_pred}). Since the resolution is low, the existing volumetric MVS method has sufficient efficiency and can be used. Second we introduce the 2D-3D feature lifting (\autoref{sec:lift}), which associates the 2D image information with 3D geometry priors. Then we propose our novel \textit{PointFlow} module (\autoref{sec:depth_refinement}) to iteratively refine the input depth map to higher resolution with improved accuracy.

\subsection{Coarse depth prediction}
\label{sec:coarse_depth_pred}
Recently, learning-based MVS~\cite{ji2017surfacenet_an-1, yao2018mvsnet_depth-1, im2018dpsnet_end-to-end} achieves state-of-the-art performance using multi-scale 3D CNNs on cost volume regularization. However, this step could be extremely memory expensive as the memory requirement is increasing cubically as the cost volume resolution grows. Taking memory and time into consideration, we use the recently proposed MVSNet~\cite{yao2018mvsnet_depth-1} to predict a relatively low-resolution cost volume.

Given the images and corresponding camera parameters, MVSNet~\cite{yao2018mvsnet_depth-1} builds a 3D cost volume upon the reference camera frustum. Then the initial depth map for reference view is regressed through multi-scale 3D CNNs and the soft argmin~\cite{knapitsch2017tanks_and} operation. In MVSNet, feature maps are downsampled to $1/4$ of the original input image in each dimension and the number of virtual depth planes are $256$ for both training and evaluation. 
On the other hand, in our coarse depth estimation network, the cost volume is constructed with feature maps of $1/8$ the size of the reference image, containing $48$ or $96$ virtual depth planes for training and evaluation, respectively. Therefore, our memory usage of this 3D feature volume is about 1/20 of that in MVSNet. 

\subsection{2D-3D feature lifting}
\label{sec:lift}
\paragraph{Image Feature Pyramid}
Learning-based image features are demonstrated to be critical to boosting up dense pixel correspondence quality ~\cite{yao2018mvsnet_depth-1, tang2018BANetDenseBundle}. In order to endow points with a larger receptive field of contextual information at multiple scales, we construct a 3-scale feature pyramid. 2D convolutional networks with stride $2$ are applied to downsample the feature map, and each last layer before the downsampling is extracted to construct the final feature pyramid $\mathbf{F}_i=[{\mathbf{F}^1_i}, {\mathbf{F}^2_i}, {\mathbf{F}^3_i}]$ for image $\mathbf{I}_i$. Similar to common MVS methods\cite{yao2018mvsnet_depth-1, im2018dpsnet_end-to-end}, feature pyramids are shared among all input images.

\paragraph{Dynamic Feature Fetching}
The point feature used in our network is compromised of the fetched multi-view image feature variance with the normalized 3D coordinates in world space $\mathbf{X}_p$. We will introduce them separately.

Image appearance features for each 3D point can be fetched from the multi-view feature maps using a differentiable unprojection given corresponding camera parameters. Note that features ${\mathbf{F}^1_i}, {\mathbf{F}^2_i}, {\mathbf{F}^3_i}$ are at different image resolutions, thus the camera intrinsic matrix should be scaled at each level of the feature maps for correct feature warping. Similar to MVSNet~\cite{yao2018mvsnet_depth-1}, we keep a variance-based cost metric, i.e. the feature variance among different views, to aggregate features warped from an arbitrary number of views. For pyramid feature at level $j$, the variance metric for $N$ views is defined as below:
\begin{equation}
\mathbf{C}^j=\frac{{{\displaystyle\sum_{i=1}^N}\left(\mathbf{F}_i^j-{\overline{\mathbf{F}^j}}\right)^2}}N, (j=1,2,3)
\end{equation}

To form the features residing at each 3D point, we do a concatenation of the fetched image feature and the normalized point coordinates:
\begin{equation}
    \mathbf{C}_p = {\rm{concat}}[\mathbf{C}_p^j, \mathbf{X}_p], (j=1,2,3)
\end{equation}
This feature augmented point $\mathbf{C}_p$ is the input to our \textit{PointFlow} module. 

As shall be seen in the next section, since we are predicting the depth residual iteratively, we update the point position $\mathbf{X}_p$ after each iteration and fetch the point feature $\mathbf{C}_p^k$ from the image feature pyramid, an operation we name as \textit{dynamic feature fetching}. Note that this step is distinct from cost-volume-based methods, by which the fetched features at each voxel are determined by the fixed space partition of the scene. In contrast, our method can fetch features from different areas of images dynamically according to the updated point position. Therefore, we can concentrate on the regions of interest in the feature maps, instead of treating them uniformly.

\subsection{\textbf{\textit{PointFlow}}}
\label{sec:depth_refinement}

Depth maps generated from \autoref{sec:coarse_depth_pred} have limited accuracy due to the low spatial resolution of 3D cost volume. We propose \textit{PointFlow}, our novel approach to iteratively refine the depth map. 

With known camera parameters, we first un-project the depth map to be a 3D point cloud. For each point, we aim to estimate its displacement to the ground truth surface along the reference camera direction by observing its neighboring points from all views, so as to push the points to \textit{flow} to the target surface. Next, we discuss the components of our module in detail.

\paragraph{Point Hypotheses Generation}
It is non-trivial to regress depth displacement of each point from the extracted image feature maps. Due to perspective transformation, the spatial context embedded in 2D feature maps cannot reflect the proximity in 3D Euclidean space.

\begin{figure}
    \centering
    \includegraphics[width=0.30\textwidth]{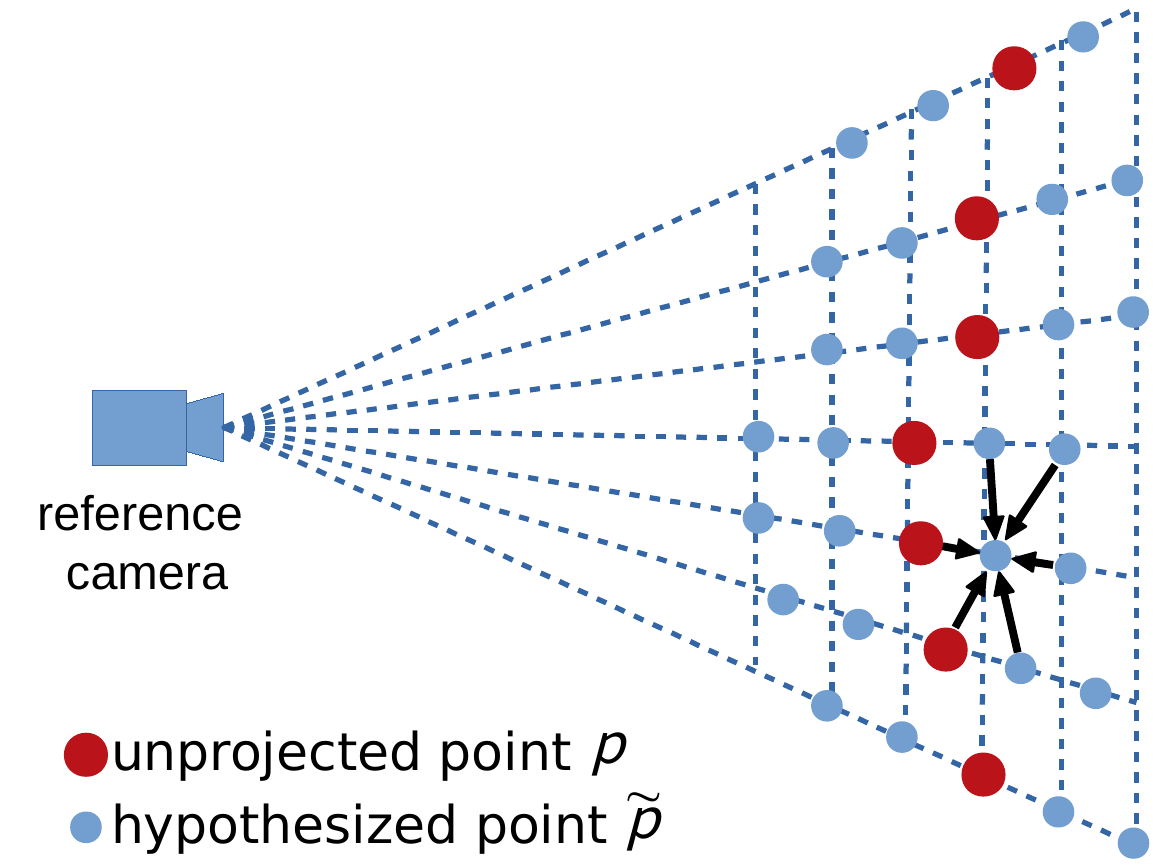}
    \caption{Illustraion of point hypotheses generation and edge construction: For each unprojected point $p$, the $2m+1$ point hypotheses $\{\mathbf{\tilde{p}}_k\}$ are generated along the reference camera direction. Directed edges are constructed between each hypothesized point and its \textsl{kNN} points for edge convolution.}
    \label{fig:edge}
\end{figure}

In order to facilitate the modeling of network, we propose to generate a sequence of \textit{point hypotheses} $\mathbf{\tilde{p}}$ with different displacement along the reference camera direction as shown in \autoref{fig:edge}. Let $\mathbf{t}$ denote the normalized reference camera direction, and $s$ denote the displacement step size. For an unprojected point $\mathbf{p}$, its hypothesized point set $\{\mathbf{\tilde{p}}_k\}$ is generated by
\begin{equation}
\mathbf{\tilde{p}}_k=\mathbf{p}+ks\mathbf{t},  \quad   k=-m, \dots, m
\label{eq:ref_points}
\end{equation}
These \textit{point hypotheses} are critical for the network to infer the displacement, for the necessary neighbourhood image feature information at different depth are gathered in these points along with spatial geometry relationship.

\paragraph{Edge Convolution}
Classical MVS methods have demonstrated that local neighborhood is important for robust depth prediction. Similarly, we take the strategy of recent work DGCNN~\cite{wang2018DynamicGraphCNN} to enrich feature aggregation between neighboring points. As shown in \autoref{fig:edge}, a directed graph is constructed on the point set using $k$ nearest neighbors (\textsl{kNN}), such that local geometric structure information could be used for the feature propagation of points.

Denote the feature augmented point cloud by ${\mathbf{C}_{\tilde{p}}}=\left\{\mathbf{C}_{\tilde{p}_1}, \dots, \mathbf{C}_{\tilde{p}_n}\right\}$, then edge convolution is defined as:
\begin{equation}
\mathbf{C}'_{\tilde{p}}=\underset{{q}\in  k\!N\!N({\tilde{p}})}\square h_\Theta\left(\mathbf{C}_{\tilde{p}},\;\mathbf{C}_{\tilde{p}}-\mathbf{C}_q\right)
\label{eq:edgeconv}
\end{equation}
where $h_\Theta$ is a learnable non-linear function parameterized by $\Theta$, and $\square$ is a channel-wise symmetric aggregation operation. There are multiple options for the symmetry operation, including max pooling, average pooling, and weighted sum. We compared max pooling and average pooling and observed similar performance after tuning hyper-parameters carefully.

\paragraph{Flow Prediction}
\begin{figure*}
\setlength{\abovecaptionskip}{0.1cm}
\setlength{\belowcaptionskip}{-0.3cm}
\centering
    \includegraphics[width=0.95\textwidth]{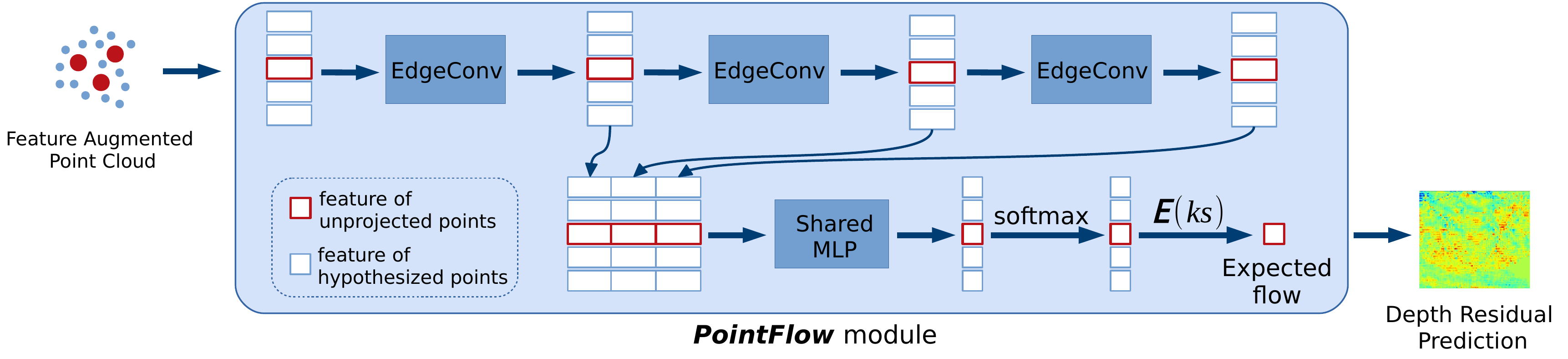}
    \caption{Network architecture of the proposed \textit{PointFlow} module.}
    \label{fig:flow}
\end{figure*}
The network architecture for flow prediction is shown in \autoref{fig:flow}. The input is a feature augmented point cloud, and the output is a depth residual map. We use three EdgeConv layers to aggregate point features at different scales of the neighborhood. Shortcut connections are used to include all the EdgeConv outputs as local point features. Finally, a shared multilayer perceptron (MLP) is used to transform the pointwise features, which outputs a probability scalar with softmax among hypothesized points of each unprojected point. The displacement of the unprojected points are predicted as the probabilistic weighted sum of the displacement among all predicted point hypotheses:
\begin{equation}
    \Delta d_{p}=\mathbf{E}(ks)=\displaystyle\sum_{k=-m}^{m} ks \times \mbox{Prob}(\mathbf{\tilde{p}}_k)
 \end{equation}

Note that this operation is differentiable. The output depth residual map is obtained by projecting the displacement back, which will be added to the initial input depth map for depth refinement.

\paragraph{Iterative Refinement with Upsampling}

Because of the flexibility of our point-based network architecture, the flow prediction can be performed iteratively, which is much harder for 3D cost-volume-based methods, because the space partitioning is fixed after the construction of cost volume.  For depth map $\mathbf{D}^{(i)}$ from coarse prediction or former residual prediction, we can first upsample it using nearest neighbor to higher spatial resolution and then perform the flow prediction to obtain $\mathbf{D}^{(i+1)}$. Moreover, we decrease the depth interval $s$ between the unprojected points and hypothesized points at each iteration, so that more accurate displacement can be predicted by capturing more detailed features from closer point hypotheses.

\subsection{Training loss}
\label{sec:training_loss}

Similar to most deep MVS networks, we treat this problem as a regression task and train the network with the $L_1$ loss, which measures the absolute difference between the predicted depth map and the groundtruth depth map. Losses for the initial depth map and iteratively refined ones are all considered:
\begin{equation}
    Loss=\sum_{i=0}^l\left(\frac{\lambda^{\left(i\right)}}{s^{\left(i\right)}}\sum_{p\in \mathbf{P}_{\rm{valid}}}{\left\|\mathbf{D}_{\rm{GT}}\left(p\right)-\mathbf{D}^{\left(i\right)}\left(p\right)\right\|}_1\right)
\label{eq:train_loss}
\end{equation}
where $\mathbf{P}_{\rm{valid}}$ represents the valid groundtruth pixel set and $l$ is the iteration number. The weight $\lambda^{\left(i\right)}$ is set to 1.0 in training.

\begin{table}[t]
\centering
\setlength{\abovecaptionskip}{0.15cm}
\footnotesize
\begin{tabular}{r|ccc}
\toprule
& Acc. (mm)    & Comp. (mm) & Overall (mm) \\
\midrule
Camp~\cite{campbell2008using_multiple}   &    0.835&    0.554&    0.695 \\
Furu~\cite{furukawa2010accurate_dense}   &0.613    &    0.941&0.777    \\
Tola~\cite{tola2012EfficientLargescaleMultiview}  &0.342    & 1.190    &0.766     \\
Gipuma~\cite{galliani2016gipuma}  &\textbf{0.283}    &0.873    & 0.578\\
SurfaceNet~\cite{ji2017surfacenet_an-1} &0.450    &1.040     & 0.745\\
MVSNet~\cite{yao2018mvsnet_depth-1}  & 0.396 & 0.527    & 0.462 \\
\midrule
Ours    & 0.361 &  0.421 & 0.391    \\
Ours-HiRes  &   0.342 & \textbf{0.411}& \textbf{0.376}    \\
\bottomrule      
\end{tabular}
\caption{Quantitative results of reconstruction quality on the DTU evaluation dataset (lower is better). }
\label{tab:res-dtu}
\end{table}

\section{Experiments}
\subsection{DTU dataset}
The DTU dataset~\cite{aanaes2016LargeScaleDataMultipleView} is a large-scale MVS dataset, which consists of 124 different scenes scanned in 7 different lighting conditions at 49 or 64 positions. The data for each scan is composed of an RGB image and corresponding camera parameters. The dataset is split into training, validation, and evaluation sets.
\subsection{Implementation details}
\paragraph{Training} We train Point-MVSNet on the DTU training dataset. For data pre-processing, we follow MVSNet~\cite{yao2018mvsnet_depth-1} to generate depth maps from the given groundtruth point clouds. During training, we set input image resolution to $640 \times 512$, and number of views to $N = 3$. The input view set is chosen with the same view selection strategy as in MVSNet (Section 4.1). For coarse prediction, we construct a 3D cost volume with $D=48$ virtual depth planes, which are uniformly sampled from $425$mm to $921$mm. For the depth refinement step, we set flow iterations $l=2$, with depth intervals being $8$mm and $4$mm, respectively. The number of nearest neighbor points $=16$. We use RMSProp of initial learning rate $0.0005$ which is decreased by 0.9 for every 2 epochs. The coarse prediction network is trained alone for 4 epochs, and then, the model is trained end-to-end for another 12 epochs. Batch size is set to 4 on 4 NVIDIA GTX 1080Ti graphics cards.

\paragraph{Evaluation} We use $D=96$ depth layers for initial depth prediction and set flow iterations $l=3$ for depth refinement. We predict the reference view depth map for each $N=5$ view set. Then we fuse all depth maps to point clouds using same post-processing provided by~\cite{yao2018mvsnet_depth-1}. We evaluate our method in two different input image resolutions: $1280\times960$ (``Ours''), and $1600\times1152$ (``Ours-HiRes'').


\begin{figure*}[t!]
\setlength{\abovecaptionskip}{0.1cm}
\setlength{\belowcaptionskip}{-0.3cm}
    \centering
    \footnotesize
    \includegraphics[width=0.8\textwidth]{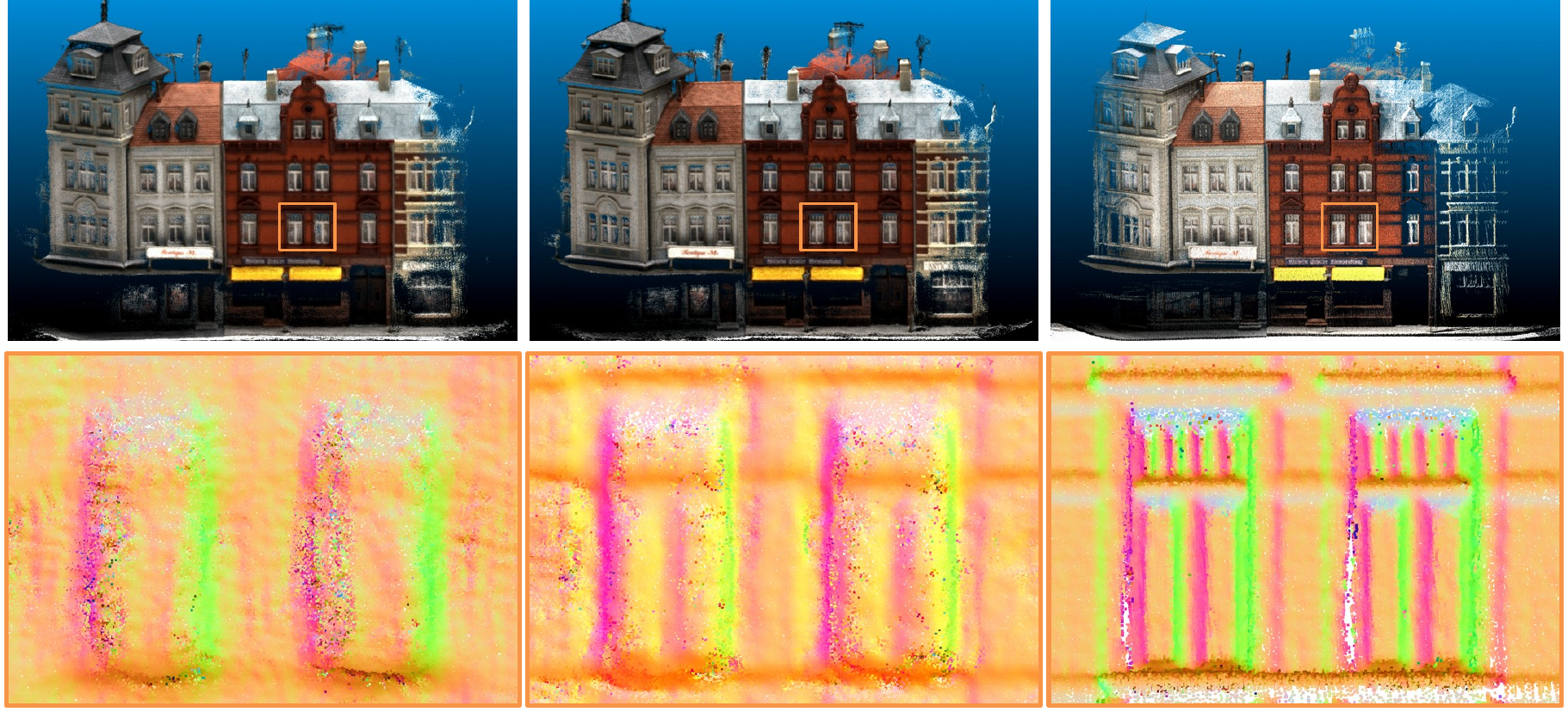}
    \begin{tabular}{x{0.25\linewidth}x{0.25\linewidth}x{0.25\linewidth}}
    MVSNet~\cite{yao2018mvsnet_depth-1} & Ours & Ground Truth\\
    \end{tabular}
    \caption{Qualitative results of \textit{scan} 9 of DTU dataset. Top: Whole point cloud. Bottom: Visualization of normals in zoomed local area. Our Point-MVSNet generates detailed point clouds with more high-frequency component than MVSNet. For fair comparison, the depth maps predicted by MVSNet are interpolated to the same resolution as our method.}
    \label{fig:scan9}
\end{figure*}
\label{sec:dtu}

\subsection{Benchmarking on DTU dataset}
\begin{figure}[t!]
\setlength{\abovecaptionskip}{0.1cm}
\setlength{\belowcaptionskip}{-0.2cm}
    \centering
    \includegraphics[width=0.35\textwidth]{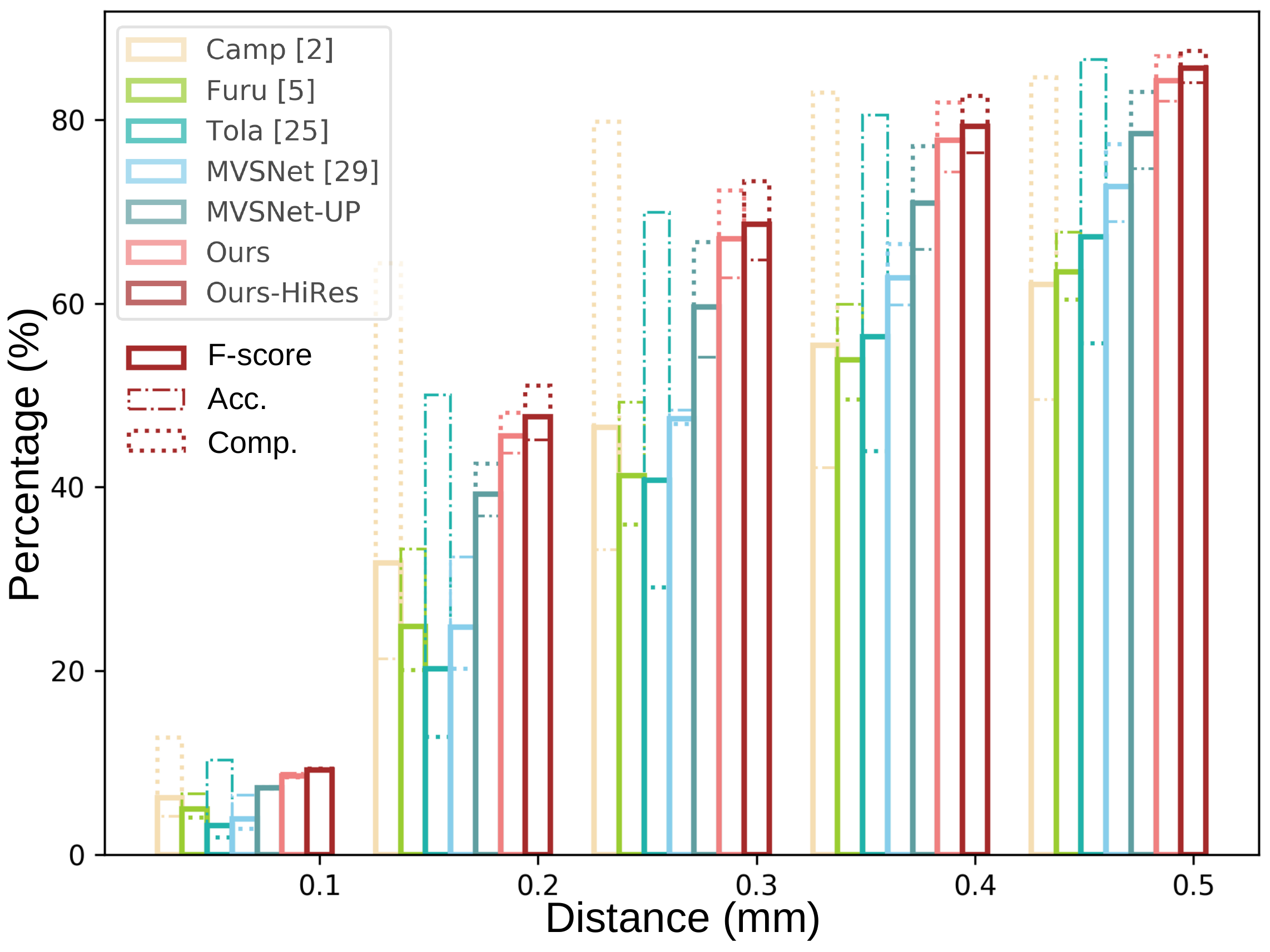}
    \caption{F-score, accuracy and completeness of different distance thresholds on the DTU evaluation dataset (higher is better).  For fair comparison, we upsample the depth map predicted by MVSNet to the same resolution as our method before depth fusion ($288\times216$ to $640\times480$).}
    \label{fig:f_curve}
\end{figure}

We evaluate the proposed method on the DTU evaluation dataset.  Quantitative results are shown in \autoref{tab:res-dtu} and \autoref{fig:f_curve}, where the accuracy and completeness are computed using the official code from the DTU dataset, and the \textit{f-score} is calculated as mentioned in~\cite{knapitsch2017tanks_and} as the measure of overall performance of accuracy and completeness. While Gipuma~\cite{galliani2016gipuma} performs the best in terms of accuracy, our Point-MVSNet outperforms start-of-the-art in both completeness and overall quality. Qualitative results are shown in \autoref{fig:scan9}. Point-MVSNet generates a more detailed point cloud compared with MVSNet. Especially in those edgy areas, our method can capture high-frequency geometric features.

\subsection{\textbf{\textit{PointFlow}} iteration}
\begin{figure*}
\setlength{\abovecaptionskip}{0.1cm}
    \centering
    \footnotesize
    \includegraphics[width=0.8\textwidth]{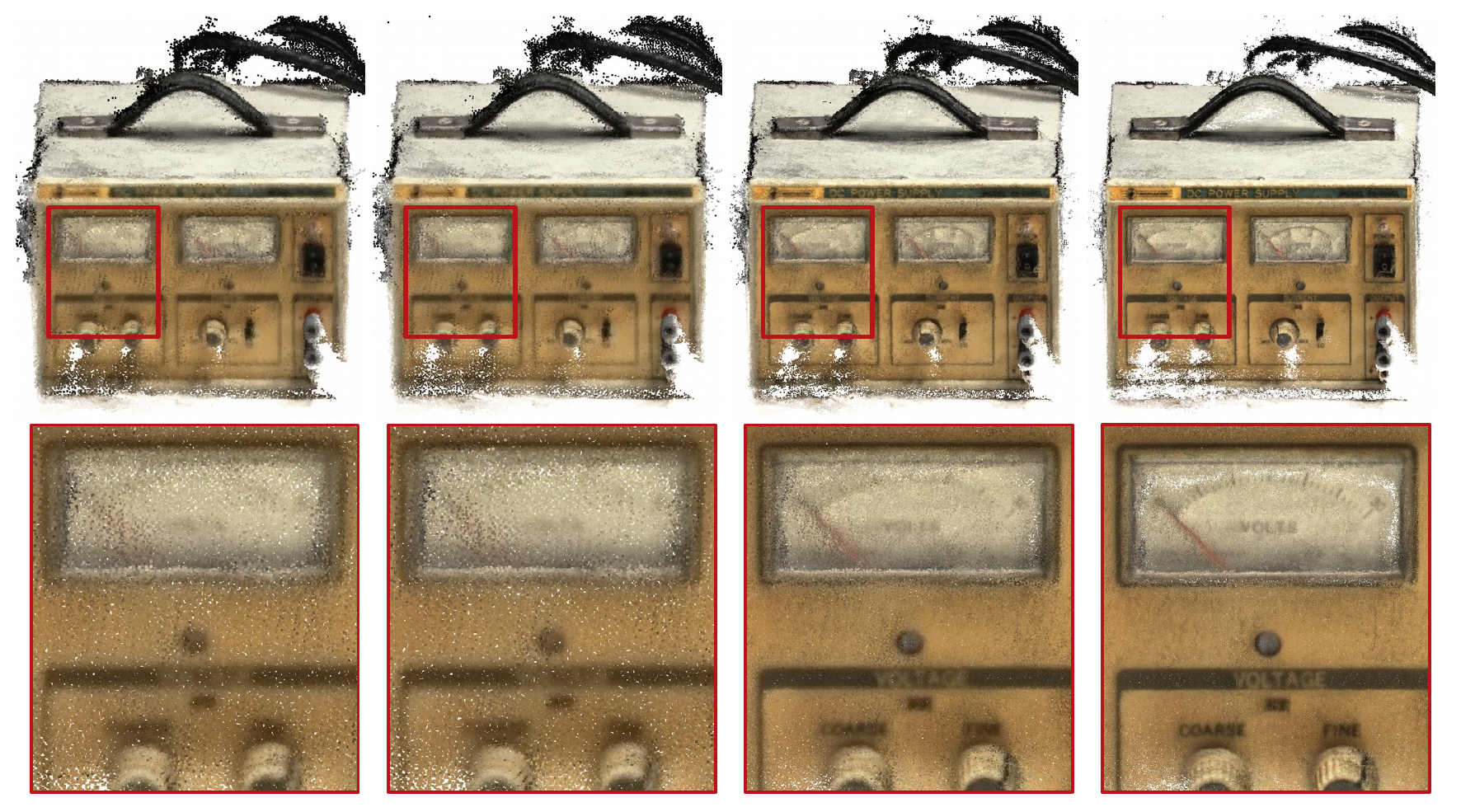}
    \begin{tabular}{x{0.18\linewidth}x{0.18\linewidth}x{0.18\linewidth}x{0.18\linewidth}}
    Initial & Iter1 & Iter2 & Iter3\\
    \end{tabular}
    \caption{Qualitative results at different flow iterations. Top: Whole point cloud. Bottom: Zoomed local area. The generated point cloud becomes denser after each iteration, and more geometry details can be captured.}
    \label{fig:iter}
\end{figure*}

\begin{table*}[]
\setlength{\abovecaptionskip}{0.15cm}
\setlength{\belowcaptionskip}{-0.3cm}
\centering
\setlength{\tabcolsep}{0.3em}
\footnotesize
\begin{tabular}{@{}c|ccc|c|cc|cc@{}}
\toprule
Iter. & Acc. (mm) & Comp. (mm) & Overall (mm) & 0.5mm \textit{f-score} & Depth Map Res. & Depth Interval (mm) & GPU Mem. (MB) & Runtime (s)  \\ \midrule
- & 0.693 & 0.758 & 0.726 & 47.95 & 160$\times$120 & 5.30 & \textbf{7219} & \textbf{0.34} \\
1 & 0.674 & 0.750 & 0.712 & 48.63 & 160$\times$120 & 5.30 & 7221 & 0.61\\
2 & 0.448 & 0.487 & 0.468 & 76.08 & 320$\times$240 & 4.00  & 7235 & 1.14\\
3 & \textbf{0.361} & \textbf{0.421} & \textbf{0.391} & \textbf{84.27} & \textbf{640$\times$480} & \textbf{0.80} & 8731 & 3.35 \\ \midrule
MVSNet\cite{yao2018mvsnet_depth-1} & 0.456 & 0.646 & 0.551 & 71.60 & 288$\times$216 & 2.65 & 10805 & 1.05\\ \bottomrule
\end{tabular}
\caption{Comparison result at different flow iterations measured by reconstruction quality and depth map resolution on the DTU evaluation set. Due to the GPU memory limitation, we decrease the resolution of MVSNet~\cite{yao2018mvsnet_depth-1} to 1152$\times$864$\times$192.}
\label{tab:flow-iter}
\end{table*}

Because of the continuity and flexibility of point representation, the refinement and densification can be performed iteratively on former predictions to give denser and more accurate predictions. While the model is trained using $l=2$ iterations, we test the model using iteration ranging from 0 to 3. For each iteration, we upsample the point cloud and decrease the depth interval of point hypotheses simultaneously, enabling the network to capture more detailed features. We compare the reconstruction quality, depth map resolution, GPU memory consumption and runtime at different iterations, along with performance reported by state-of-the-art methods in \autoref{tab:flow-iter}. The reconstruction quality improves significantly with multiple iterations, which verifies the effectiveness of our methods. Note that our method already outperforms the state-of-the-art after the second iteration. Qualitative results are shown in \autoref{fig:iter}. 

\subsection{Ablation study}

In this section we provide ablation experiments and quantitative analysis to evaluate the strengths and limitations of the key components in our framework. For all the following studies, experiments are performed and evaluated on the DTU dataset, and both \textsl{accuracy} and \textsl{completeness} are used to measure the reconstruction quality. We set the iteration number to $l=2$, and all other experiment settings are the same as \autoref{sec:dtu}.

\paragraph{Edge Convolution}
By replacing the edge convolution (~\ref{eq:edgeconv}) with geometry-unaware feature aggregation:
\begin{equation}
\mathbf{C}'_{\tilde{p}}=\underset{{q}\in  k\!N\!N({\tilde{p}})}\square h_\Theta\left(\mathbf{C}_q\right),
\end{equation}
where the features of neighbor points are treated equally with no regard for their geometric relationship to the centroid point, the reconstruction quality drops significantly as shown in \autoref{tab:ablation}, which illustrates the importance of local neighborhood relationship information (captured by $\mathbf{C}_{\tilde{p}}-\mathbf{C}_{q}$) for feature aggregation. 

\paragraph{Euclidean Nearest Neighbour}
In this part, we construct the directed graph $\mathcal G$ using points belonging to adjacent pixels in the reference image, instead of searching the $k$-NN points, which leads to decreased reconstruction quality. The reason is that, for images of 3D scenes, near-by pixels may correspond to distant objects due to occlusion. Therefore, using neighboring points in the image space may aggregate irrelevant features for depth residual prediction, leading to descending performance.

\paragraph{Feature Pyramid}
In this part, point cloud only fetches features from the last layer of the feature map, instead of from the whole feature pyramid. As shown in \autoref{tab:ablation}, in contrast to the relatively stable performance for changing edge convolution strategies as discussed above, the drop will be significant in the absence of the other two components, which demonstrates the effectiveness of the leveraging context information at different scales for feature fetching.

\subsection{Reliance on initial depth maps}
Our method uses state-of-the-art approaches to get a coarse depth map prediction, which is then iteratively refined by predicting depth residuals. We found that our approach is robust to noisy initial depth estimation in a certain range through the following experiments.  We added Gaussian noise of different scales to the initial depth map and evaluated the reconstruction error. \autoref{fig:sensitive} shows that the error increases slowly and is smaller than MVSNet within 6mm noise. 

\begin{table}[t]
\setlength{\abovecaptionskip}{0.15cm}
\centering
\footnotesize
\begin{tabular}{cccccc}
\toprule
EDGE & EUCNN & PYR & Acc. (mm) & Comp. (mm) \\
\midrule
\checkmark & \checkmark & \checkmark &   \textbf{0.448}   &  \textbf{0.487} \\
\checkmark & \checkmark & \red{\ding{53}} &  0.455& 0.489 \\
\checkmark & \red{\ding{53}} & \checkmark & 0.455 & 0.492 \\
\red{\ding{53}}& \checkmark & \checkmark & 0.501 & 0.518\\
\checkmark & \red{\ding{53}} & \red{\ding{53}} & 0.475 & 0.504 \\
\red{\ding{53}} & \checkmark & \red{\ding{53}} &0.574 & 0.565 \\
\red{\ding{53}}& \red{\ding{53}} & \checkmark & 0.529 & 0.532\\
\bottomrule
\end{tabular}
\caption{Ablation study on network architectures on the DTU evaluation dataset, which demonstrates the effectiveness of different components. EDGE denotes edge convolution, EUCNN denotes grouping by nearest neighbour points in Euclidean distance, and PYR denotes the usage of image feature pyramid.}
\label{tab:ablation}
\end{table}


\begin{figure}[t]
\setlength{\abovecaptionskip}{0.0cm}
\begin{center}
  \includegraphics[width=0.7\linewidth]{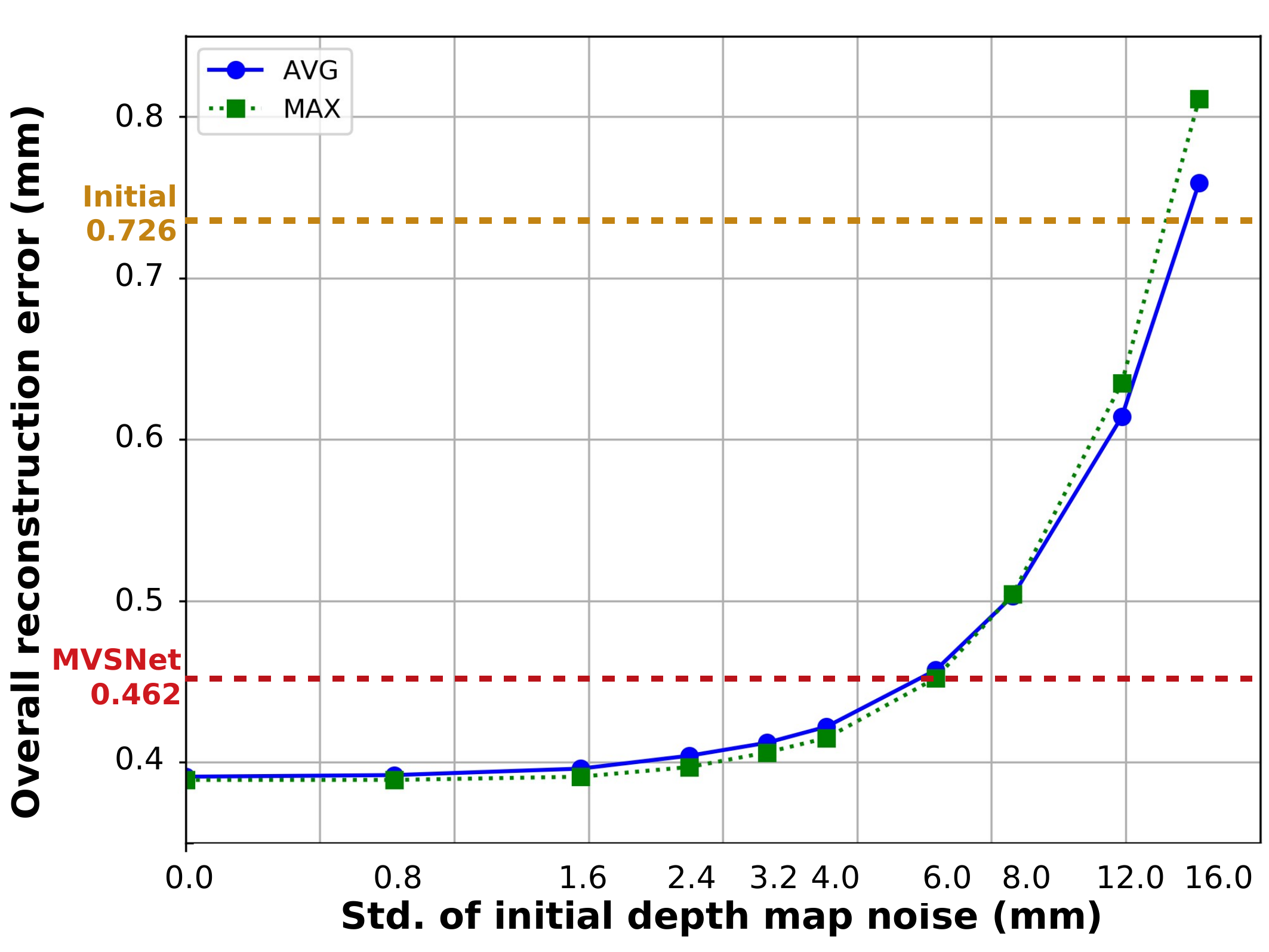}
\end{center}
  \caption{{Reconstruction error w.r.t. initial depth map noise. AVG denotes average pooling, MAX denotes max pooling.}}
\label{fig:sensitive}
\end{figure}
    


\begin{table}[]
\setlength{\abovecaptionskip}{0.15cm}
    \footnotesize
    \centering
    \begin{tabular}{cccc}
    \toprule
         & Acc. (mm) & Comp. (mm) & Overall (mm) \\
    \midrule 
    PU-Net~\cite{yu2018punet}& 1.220 & 0.667&0.943 \\
    Ours &  \textbf{0.361} &  \textbf{0.421} & \textbf{0.391} \\
    \bottomrule
    \end{tabular}
    \caption{Comparison of reconstruction quality on the DTU evaluation dataset with PU-Net~\cite{yu2018punet}. }
    \label{tab:punet}
\end{table}

\subsection{Comparison to point cloud upsampling}
Our work can also be considered as a data-driven point cloud upsampling method with assisting information from reference views. Therefore, we compare our method with PU-Net~\cite{yu2018punet}, where multi-level features are extracted from the coarse point cloud to reconstruct an upsampled point cloud. 

We use the same coarse depth prediction network as in our model, and train PU-Net to upsample the coarse point cloud.  We use the same joint loss as mentioned in their paper, which consists of two losses --- the Earth Mover's distance (EMD)~\cite{fan2017point} loss between the predicted point cloud and the reference groundtruth point cloud and a repulsion loss. For evaluation, the PU-Net is applied on the coarse predicted point cloud twice to generate a denser point cloud with $16$ times more points. Quantitative result is shown in \autoref{tab:punet}. Our Point-MVSNet can generate a more accurate point cloud from the coarse one by inducing \textit{flow} for each point from observation of context information in multi-view images.

\subsection{Foveated depth inference}
\begin{figure}
\centering
    \includegraphics[width=0.7\linewidth]{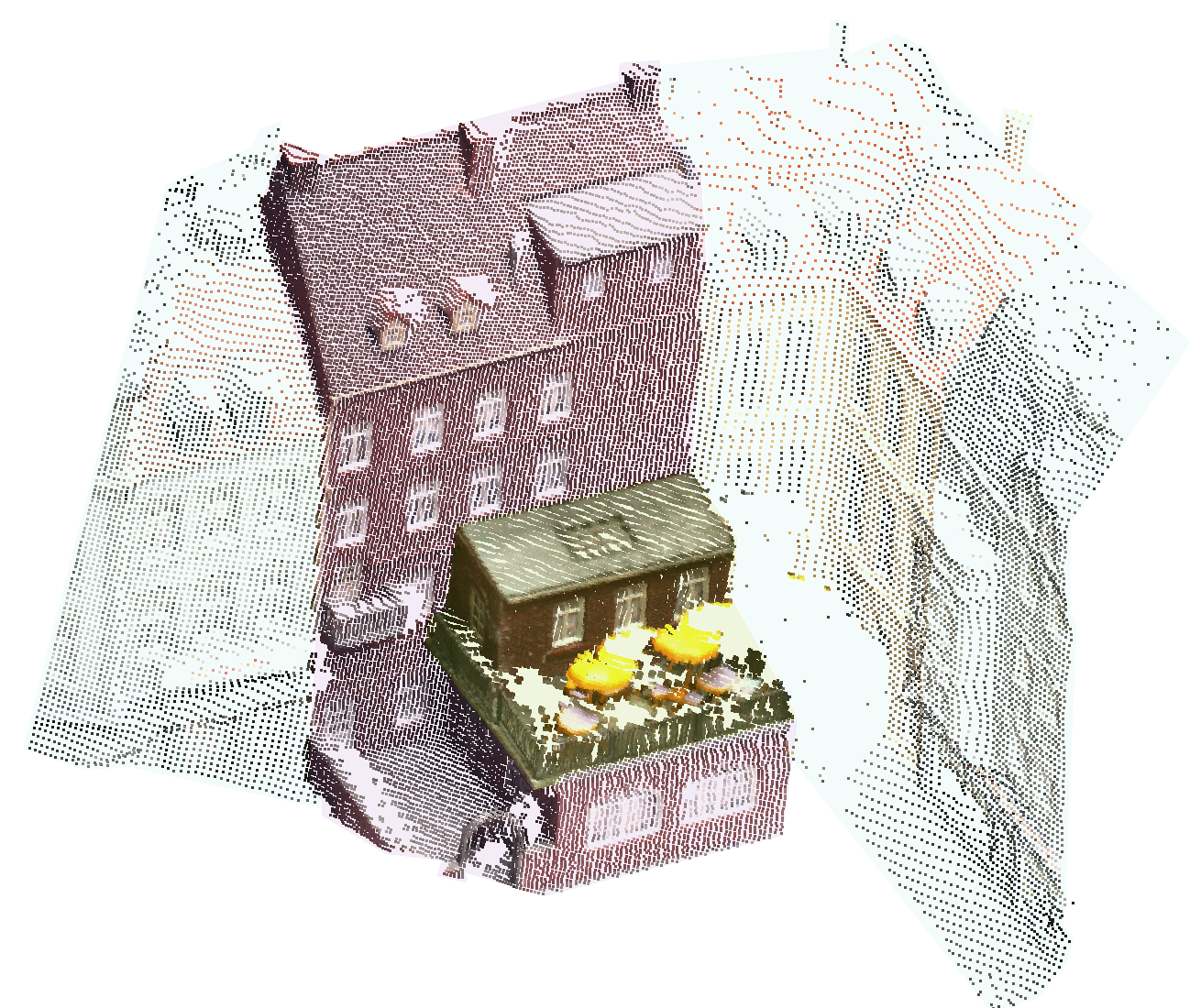}
    \caption{Illustration of foveated depth inference with our proposed method. Different point density levels are denoted by different colors: Gray for sparsest, Brown for intermediate, Green for densest.}
    \label{fig:adapt_pt}
\end{figure}

The point-based network architecture enables us to process an arbitrary number of points. Therefore, instead of upsampling and refining the whole depth map, we can choose to only infer the depth in the region of interest (ROI) based on the input image or the predicted coarse depth map. As shown in \autoref{fig:adapt_pt}, we generate a point cloud of three different density levels by only upsampling and refining the ROI in the previous stage.

\subsection{Generalizability of the \textbf{\textit{PointFlow}} Module}
In order to evaluate the generalizability of our \textit{PointFlow} module, we test it on the Tanks and Temples intermediate dataset~\cite{knapitsch2017tanks_and}, which is a large outdoor dataset captured in complex environments. We first generate coarse depth maps using MVSNet~\cite{yao2018mvsnet_depth-1}, and then apply our \textit{PointFlow} module to refine them. The \textit{f-score} increases from $43.48$ to $48.27$ (larger is better) and the rank rises from $13.12$ to $7.25$ (lower is better, date: Mar. 22, 2019). Reconstructed point clouds are shown in supplementary materials.

\section{Conclusion}
We present a novel point-based architecture for high-resolution multi-view stereo reconstruction. Instead of building a high-resolution cost volume, our proposed Point-MVSNet processes the scene as a point cloud directly, which reduces unnecessary computation and preserves the spatial continuity. Experiments show that Point-MVSNet is able to produce high-quality reconstruction point clouds on benchmarks. Additionally, Point-MVSNet is applicable to foveated depth inference to greatly reducing computation, which cannot be easily implemented for cost-volume-based methods.

\section*{Acknowledgement}
The authors gratefully acknowledge the support of an NSF grant IIS-1764078, gifts from Qualcomm, Adobe and support from DMAI corporations.



{\small
\bibliographystyle{ieee_fullname}
\bibliography{main2.bib}

\begin{thebibliography}{10}\itemsep=-1pt

\bibitem{aanaes2016LargeScaleDataMultipleView}
Henrik {Aan{\ae}s}, Rasmus~Ramsb\O{}l Jensen, George Vogiatzis, Engin Tola, and
  Anders~Bjorholm Dahl.
\newblock Large-{{Scale Data}} for {{Multiple}}-{{View Stereopsis}}.
\newblock {\em Int. J. Comput. Vision}, 120(2):153--168, Nov. 2016.

\bibitem{campbell2008using_multiple}
Neill D.~F. Campbell, George Vogiatzis, Carlos Hern\'andez, and Roberto
  Cipolla.
\newblock Using {{Multiple Hypotheses}} to {{Improve Depth}}-{{Maps}} for
  {{Multi}}-{{View Stereo}}.
\newblock In David Forsyth, Philip Torr, and Andrew Zisserman, editors, {\em
  Computer {{Vision}} \textendash{} {{ECCV}} 2008}, volume 5302, pages
  766--779. 2008.

\bibitem{esteban2004silhouette}
Carlos~Hern{\'a}ndez Esteban and Francis Schmitt.
\newblock Silhouette and stereo fusion for 3d object modeling.
\newblock {\em Computer Vision and Image Understanding}, 96(3):367--392, 2004.

\bibitem{fan2017point}
Haoqiang Fan, Hao Su, and Leonidas~J Guibas.
\newblock A point set generation network for 3d object reconstruction from a
  single image.
\newblock In {\em Proceedings of the IEEE conference on computer vision and
  pattern recognition}, pages 605--613, 2017.

\bibitem{furukawa2010accurate_dense}
Y. Furukawa and J. Ponce.
\newblock Accurate, {{Dense}}, and {{Robust Multiview Stereopsis}}.
\newblock {\em IEEE Transactions on Pattern Analysis and Machine Intelligence},
  32(8):1362--1376, Aug. 2010.

\bibitem{galliani2016gipuma}
Silvano Galliani, Katrin Lasinger, and Konrad Schindler.
\newblock Gipuma: Massively parallel multi-view stereo reconstruction.
\newblock {\em Publikationen der Deutschen Gesellschaft f{\"u}r
  Photogrammetrie, Fernerkundung und Geoinformation e. V}, 25:361--369, 2016.

\bibitem{han2015matchnet}
Xufeng Han, Thomas Leung, Yangqing Jia, Rahul Sukthankar, and Alexander~C Berg.
\newblock Matchnet: Unifying feature and metric learning for patch-based
  matching.
\newblock In {\em Proceedings of the IEEE Conference on Computer Vision and
  Pattern Recognition}, pages 3279--3286, 2015.

\bibitem{he2016resnet}
Kaiming He, Xiangyu Zhang, Shaoqing Ren, and Jian Sun.
\newblock Deep residual learning for image recognition.
\newblock In {\em Proceedings of the IEEE conference on computer vision and
  pattern recognition}, pages 770--778, 2016.

\bibitem{hornung2006hierarchical}
Alexander Hornung and Leif Kobbelt.
\newblock Hierarchical volumetric multi-view stereo reconstruction of manifold
  surfaces based on dual graph embedding.
\newblock In {\em 2006 IEEE Computer Society Conference on Computer Vision and
  Pattern Recognition (CVPR'06)}, volume~1, pages 503--510. IEEE, 2006.

\bibitem{huang2018deepmvs_learning}
Po-Han Huang, Kevin Matzen, Johannes Kopf, Narendra Ahuja, and Jia-Bin Huang.
\newblock {{DeepMVS}}: {{Learning Multi}}-view {{Stereopsis}}.
\newblock {\em arXiv:1804.00650 [cs]}, Apr. 2018.

\bibitem{im2018dpsnet_end-to-end}
Sunghoon Im, Hae-Gon Jeon, Stephen Lin, and In~So Kweon.
\newblock {{DPSNet}}: {{End}}-to-end {{Deep Plane Sweep Stereo}}.
\newblock Sept. 2018.

\bibitem{ji2017surfacenet_an-1}
Mengqi Ji, Juergen Gall, Haitian Zheng, Yebin Liu, and Lu Fang.
\newblock {{SurfaceNet}}: {{An End}}-to-end {{3D Neural Network}} for
  {{Multiview Stereopsis}}.
\newblock {\em arXiv:1708.01749 [cs]}, Aug. 2017.

\bibitem{kar2017lsm}
Abhishek Kar, Christian H{\"a}ne, and Jitendra Malik.
\newblock Learning a multi-view stereo machine.
\newblock In {\em Advances in neural information processing systems}, pages
  365--376, 2017.

\bibitem{kendall2017gcnet}
Alex Kendall, Hayk Martirosyan, Saumitro Dasgupta, Peter Henry, Ryan Kennedy,
  Abraham Bachrach, and Adam Bry.
\newblock End-to-end learning of geometry and context for deep stereo
  regression.
\newblock In {\em Proceedings of the IEEE International Conference on Computer
  Vision}, pages 66--75, 2017.

\bibitem{knapitsch2017tanks_and}
Arno Knapitsch, Jaesik Park, Qian-Yi Zhou, and Vladlen Koltun.
\newblock Tanks and {{Temples}}: {{Benchmarking Large}}-scale {{Scene
  Reconstruction}}.
\newblock {\em ACM Trans. Graph.}, 36(4):78:1--78:13, July 2017.

\bibitem{knobelreiter2017cnncrf}
Patrick Knobelreiter, Christian Reinbacher, Alexander Shekhovtsov, and Thomas
  Pock.
\newblock End-to-end training of hybrid cnn-crf models for stereo.
\newblock In {\em Proceedings of the IEEE Conference on Computer Vision and
  Pattern Recognition}, pages 2339--2348, 2017.

\bibitem{lhuillier2005quasi}
Maxime Lhuillier and Long Quan.
\newblock A quasi-dense approach to surface reconstruction from uncalibrated
  images.
\newblock {\em IEEE transactions on pattern analysis and machine intelligence},
  27(3):418--433, 2005.

\bibitem{owens2013shape_anchors}
A. Owens, J. Xiao, A. Torralba, and W. Freeman.
\newblock Shape {{Anchors}} for {{Data}}-{{Driven Multi}}-view
  {{Reconstruction}}.
\newblock In {\em 2013 {{IEEE International Conference}} on {{Computer
  Vision}}}, pages 33--40, Dec. 2013.

\bibitem{qi2016PointNetDeepLearning}
Charles~R. Qi, Hao Su, Kaichun Mo, and Leonidas~J. Guibas.
\newblock {{PointNet}}: {{Deep Learning}} on {{Point Sets}} for {{3D
  Classification}} and {{Segmentation}}.
\newblock {\em arXiv:1612.00593 [cs]}, Dec. 2016.

\bibitem{qi2017PointNetDeepHierarchical}
Charles~R. Qi, Li Yi, Hao Su, and Leonidas~J. Guibas.
\newblock {{PointNet}}++: {{Deep Hierarchical Feature Learning}} on {{Point
  Sets}} in a {{Metric Space}}.
\newblock {\em arXiv:1706.02413 [cs]}, June 2017.

\bibitem{riegler2017octnet}
Gernot Riegler, Ali Osman~Ulusoy, and Andreas Geiger.
\newblock Octnet: Learning deep 3d representations at high resolutions.
\newblock In {\em Proceedings of the IEEE Conference on Computer Vision and
  Pattern Recognition}, pages 3577--3586, 2017.

\bibitem{seki2017sgm}
Akihito Seki and Marc Pollefeys.
\newblock Sgm-nets: Semi-global matching with neural networks.
\newblock In {\em Proceedings of the IEEE Conference on Computer Vision and
  Pattern Recognition}, pages 231--240, 2017.

\bibitem{tang2018BANetDenseBundle}
Chengzhou Tang and Ping Tan.
\newblock {{BA}}-{{Net}}: {{Dense Bundle Adjustment Network}}.
\newblock June 2018.

\bibitem{tatarchenko2017octreegen}
Maxim Tatarchenko, Alexey Dosovitskiy, and Thomas Brox.
\newblock Octree generating networks: Efficient convolutional architectures for
  high-resolution 3d outputs.
\newblock In {\em Proceedings of the IEEE International Conference on Computer
  Vision}, pages 2088--2096, 2017.

\bibitem{tola2012EfficientLargescaleMultiview}
Engin Tola, Christoph Strecha, and Pascal Fua.
\newblock Efficient {{Large}}-scale {{Multi}}-view {{Stereo}} for {{Ultra
  High}}-resolution {{Image Sets}}.
\newblock {\em Mach. Vision Appl.}, 23(5):903--920, Sept. 2012.

\bibitem{vogiatzis2007multiview}
George Vogiatzis, Carlos~Hern{\'a}ndez Esteban, Philip~HS Torr, and Roberto
  Cipolla.
\newblock Multiview stereo via volumetric graph-cuts and occlusion robust
  photo-consistency.
\newblock {\em IEEE Transactions on Pattern Analysis and Machine Intelligence},
  29(12):2241--2246, 2007.

\bibitem{Wang-2017-OCNN}
Peng-Shuai Wang, Yang Liu, Yu-Xiao Guo, Chun-Yu Sun, and Xin Tong.
\newblock {O-CNN: Octree-based Convolutional Neural Networks for 3D Shape
  Analysis}.
\newblock {\em ACM Transactions on Graphics (SIGGRAPH)}, 36(4), 2017.

\bibitem{wang2018DynamicGraphCNN}
Yue Wang, Yongbin Sun, Ziwei Liu, Sanjay~E. Sarma, Michael~M. Bronstein, and
  Justin~M. Solomon.
\newblock Dynamic {{Graph CNN}} for {{Learning}} on {{Point Clouds}}.
\newblock {\em arXiv:1801.07829 [cs]}, Jan. 2018.

\bibitem{yao2018mvsnet_depth-1}
Yao Yao, Zixin Luo, Shiwei Li, Tian Fang, and Long Quan.
\newblock {{MVSNet}}: {{Depth Inference}} for {{Unstructured Multi}}-view
  {{Stereo}}.
\newblock {\em arXiv:1804.02505 [cs]}, Apr. 2018.

\bibitem{yu2018punet}
Lequan Yu, Xianzhi Li, Chi-Wing Fu, Daniel Cohen-Or, and Pheng-Ann Heng.
\newblock Pu-net: Point cloud upsampling network.
\newblock In {\em Proceedings of IEEE Conference on Computer Vision and Pattern
  Recognition (CVPR)}, 2018.

\bibitem{zaharescu2007transformesh}
Andrei Zaharescu, Edmond Boyer, and Radu Horaud.
\newblock Transformesh: a topology-adaptive mesh-based approach to surface
  evolution.
\newblock In {\em Asian Conference on Computer Vision}, pages 166--175.
  Springer, 2007.

\end{thebibliography}
}

\clearpage
\noindent {\LARGE {\bf Supplementary Materials }}
\section{Additional Ablation Study}
\subsection{Number of Point Hypotheses}
In this section, we conduct an ablation study to verify the influence of the number of point hypotheses. In the main paper, we choose $m=2$ for both the training and evaluation. We change to $m=1$ and $m=3$, and conduct the evaluation on the DTU evaluation set~\cite{aanaes2016LargeScaleDataMultipleView}. \autoref{tab:point_hypo} shows the comparison result. Our proposed algorithm achieves best reconstruction quality in terms of completeness and overall quality when the number of point hypotheses is $m=2$.

\begin{table}[htbp]
    \centering
    \small
    \begin{tabular}{c|ccc}
    \toprule
    Point Hypotheses & Acc.(mm) & Comp.(mm) & Overall(mm) \\
    \hline
    1 & \textbf{0.442} & 0.515 & 0.479\\
    2 & 0.448 & \textbf{0.487} & \textbf{0.468}\\
    3 & 0.468 & 0.499 & 0.484\\
    \bottomrule
    \end{tabular}
    \caption{Ablation study of different number of point hypotheses $m$ on the DTU evaluation set~\cite{aanaes2016LargeScaleDataMultipleView}. (The model is trained with $m=2$.)}
    \label{tab:point_hypo}
\end{table}

\subsection{Number of Views}
In this section, we study the influence of the number of input views $N$. Utilizing a variance-based cost metric, our Point-MVSNet can process an arbitrary number of input views. Although the model is trained using $N=3$, we can evaluate the model using either $N=2, 3, 5$ on the DTU evaluation set~\cite{aanaes2016LargeScaleDataMultipleView}. \autoref{tab:num_view} demonstrates that the reconstruction quality improves with an increasing number of input views, which is consistent with common knowledge of MVS reconstruction.
\begin{table}[htbp]
 \centering
 \small
    \begin{tabular}{c|ccc}
    \toprule
    Number of Views & Acc.(mm) & Comp. (mm) & Overall(mm) \\
    \hline
    2 & 0.462 & 0.604 & 0.533 \\
    3 & 0.448 & 0.507 & 0.478\\
    5 & \textbf{0.448} & \textbf{0.487} & \textbf{0.468} \\
    \bottomrule
    \end{tabular}
    \caption{Ablation study on different number of input views $N$ on the DTU evaluation set~\cite{aanaes2016LargeScaleDataMultipleView}. (The model is trained with $N=3$)}
    \label{tab:num_view}
\end{table}

\section{Memory, runtime and overhead of \textsl{kNN}}
\autoref{tab:mem-comp} compares our memory usage and running speed against MVSNet. Our method is able to predict different resolutions of depth maps at different speed by changing the iterations. Na\"ive \textsl{kNN} of point cloud of $N$ points can be memory-consuming with $O(N^2)$ complexity. However, we notice the \textsl{kNN} of a point tend to come from its nearby 2D pixels in the depth map. By leveraging this fact and taking the hypothetical points into consideration, we restrict the \textsl{kNN} search of each point from the whole point cloud to its $k\times k \times (2m+1)$ neighborhood. Furthermore, we parallel the distance computation by using a fixed weight 3D kernel. 
 
\begin{table}[ht]
\centering
\small
\begin{threeparttable}
\begin{tabular}{c|cccc}
\hline
Iter. & \begin{tabular}[c]{@{}c@{}}Overall Err. \\ (mm)\end{tabular} & Resolution & \begin{tabular}[c]{@{}c@{}}GPU Mem.\\ (MB)\end{tabular} & \begin{tabular}[c]{@{}c@{}}Runtime\\ (s)\end{tabular} \\ \hline
0 & 0.726 & 160$\times$120 & 7219 & 0.34 \\
1 & 0.712 & 160$\times$120 & 7221 & 0.61 \\
2 & 0.468 & 320$\times$240 & 7235 & 1.14 \\
2\tnote{$\dagger$} & 0.474 & 320$\times$240 & 7233 & 0.97 \\
3 & 0.391 & 640$\times$480 & 8731 & 3.35 \\ \hline
MVSNet & 0.551 & 288$\times$216 & 10805 & 1.05 \\ \hline
\end{tabular}
\caption{Comparison of memory consumption and runtime. kNN is used for grouping, where all iterations adopt Euclidean distance, except for the iteration that is indicated by~$\dagger$, which uses pixel neighbor.}
\label{tab:mem-comp}
\end{threeparttable}
\end{table}

\section{Post-processing}
In this section, we describe the post-processing procedure in details. Similar to MVSNet~\cite{yao2018mvsnet_depth-1}, our post-processing is composed of three steps: photometric filtering, geometric consistency filtering, and depth fusion.

For photometric filtering, we use predicted probability of the most likely depth layer as the confidence metric and filter out points whose confidence is below a threshold. The filtering threshold is set to 0.5 and 0.2 for coarse and our PointFlow stage, respectively. For geometric consistency, we calculate the discrepancy of predicted depths among multi-view predictions through reverse-projection. Points with discrepancy larger than 0.12mm are discarded. For depth fusion, we take average value of all reprojected depths of each point in visible views as the final depth prediction and produce the 3D point cloud.

\section{Reconstruction Results}
\begin{figure*}
    \centering
    \includegraphics[width=0.95\textwidth]{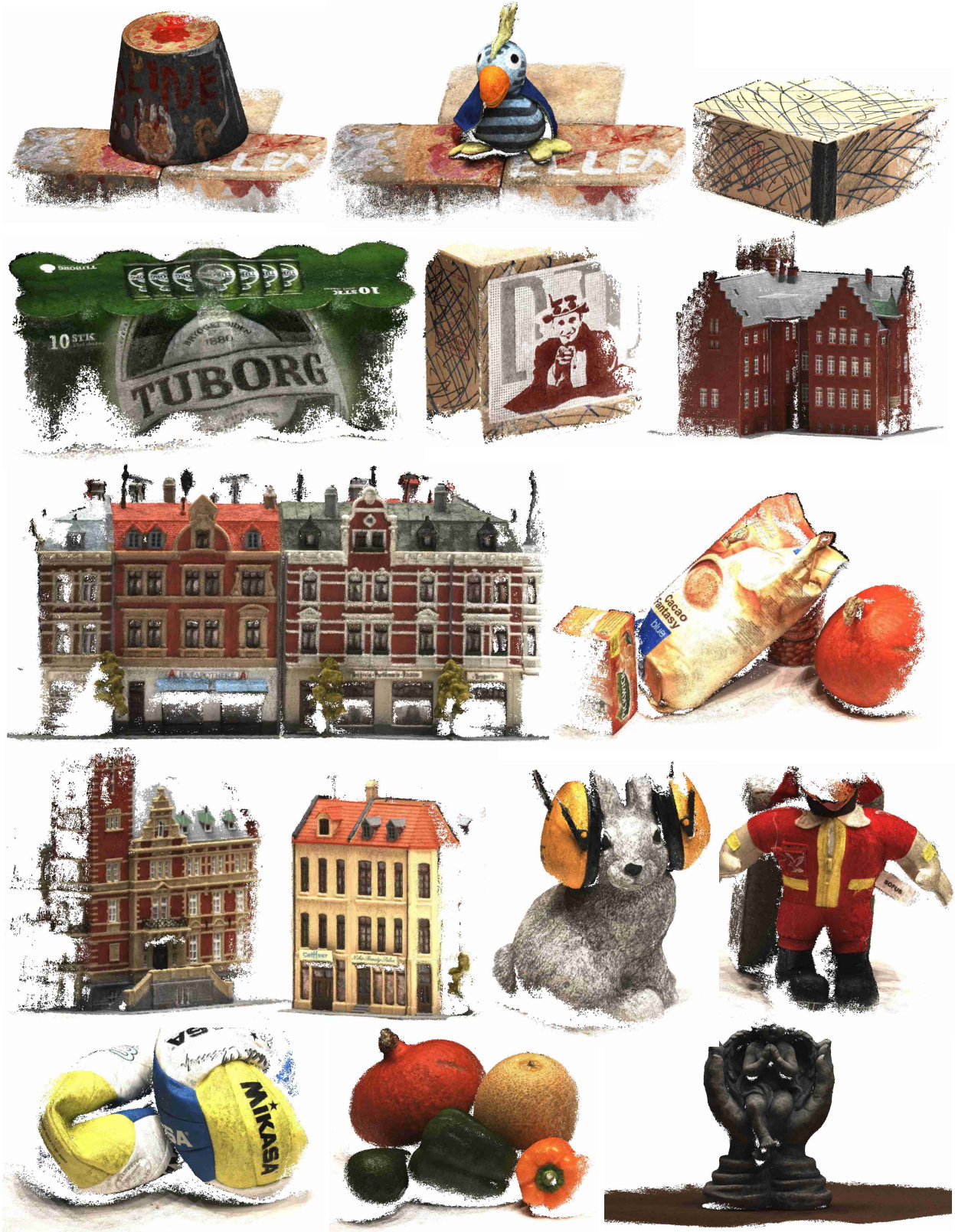}
    \caption{Reconstruction results on the DTU evaluation set~\cite{aanaes2016LargeScaleDataMultipleView}.}
    \label{fig:dtu}
\end{figure*}

\begin{figure*}
    \centering
    \includegraphics[width=0.95\textwidth]{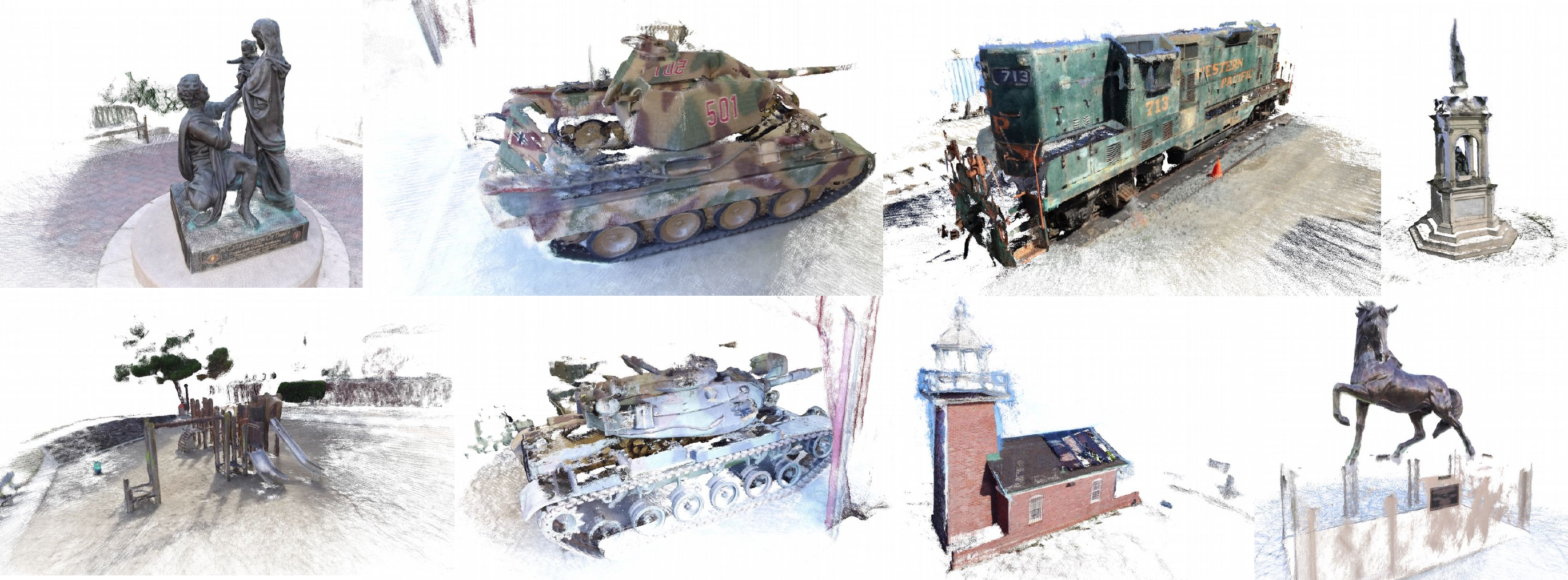}
    \caption{Reconstruction results on the intermediate set of Tanks and Temples~\cite{knapitsch2017tanks_and}.}
    \label{fig:tanks}
\end{figure*}

This section shows the reconstruction results of DTU dataset~\cite{aanaes2016LargeScaleDataMultipleView} and Tanks and Temples dataset~\cite{knapitsch2017tanks_and} in \autoref{fig:dtu} and \autoref{fig:tanks} respectively. Point-MVSNet is able to reconstruct dense and accurate point clouds for all scenes.

\end{document}